\documentclass[letterpaper, 10 pt, conference]{ieeeconf}  

\IEEEoverridecommandlockouts                              

\usepackage{cite}
\usepackage{url}
\usepackage{multirow}
\usepackage{subcaption}
\usepackage{float}
\usepackage{dblfloatfix} 
\usepackage{placeins} 
\usepackage{amsmath,amssymb,amsfonts}
\usepackage{algorithmic}
\usepackage{graphicx}
\usepackage{booktabs}
\usepackage{caption}
\usepackage{pifont}
\usepackage{tabularx}
\usepackage{textcomp}
\usepackage{xcolor}
\def\BibTeX{{\rm B\kern-.05em{\sc i\kern-.025em b}\kern-.08em
    T\kern-.1667em\lower.7ex\hbox{E}\kern-.125emX}}
\begin{document}

\title{Leveraging RAG-LLMs for Urban Mobility Simulation and Analysis
\author{Yue Ding, Conor McCarthy, Kevin O'Shea and Mingming Liu  \thanks{Y. Ding is with the Centre for Research Training in Machine Learning (ML-Labs) at Dublin City University. C. McCarthy, K. O'Shea and M. Liu are with the Insight Centre for Data Analytics and the School of Electronic Engineering, Dublin City University, Dublin, Ireland. This work has emanated from research supported in part by Insight Centre for Data Analytics at Dublin City University
under Grant Number \textit{SFI/12/RC/2289\_P2} and by Éireann – Research Ireland through the Research Ireland Centre for Research Training in Machine Learning (18/CRT/6183). \textit{Corresponding author: Mingming Liu. Email: {\tt mingming.liu@dcu.ie}.}}}}

\newcommand{\xmark}{\ding{55}}


\maketitle

\begin{abstract}
With the rise of smart mobility and shared e-mobility services, numerous advanced technologies have been applied to this field. Cloud-based traffic simulation solutions have flourished, offering increasingly realistic representations of the evolving mobility landscape. LLMs have emerged as pioneering tools, providing robust support for various applications, including intelligent decision-making, user interaction, and real-time traffic analysis. As user demand for e-mobility continues to grow, delivering comprehensive end-to-end solutions has become crucial. In this paper, we present a cloud-based, LLM-powered shared e-mobility platform, integrated with a mobile application for personalized route recommendations. The optimization module is evaluated based on travel time and cost across different traffic scenarios. Additionally, the LLM-powered RAG framework is evaluated at the schema level for different users, using various evaluation methods. Schema-level RAG with XiYanSQL achieves an average execution accuracy of 0.81 on system operator queries and 0.98 on user queries. 

\end{abstract}

\begin{keywords}
Smart Mobility, Shared E-Mobility, Traffic Simulation, Route Optimization, RAG, Cloud-Based Platform.
\vspace{-0.2in}
\end{keywords}

\section{Introduction}
Recent advances in smart mobility, information technologies, and data analytics have enabled the development of sustainable transportation solutions \cite{factorsinfluencingemobility}. Building on these, the Mobility as a Service (MaaS) paradigm integrates public transit, shared mobility, and private vehicles into a unified, user-centric platform. Within MaaS, shared e-mobility services like e-cars, e-bikes, and e-scooters provide flexible, on-demand alternatives to private vehicles, promoting sustainable urban travel \cite{Hasselwander2023}. With the growing deployment of eHubs, which are clusters of shared mobility services such as e-cars and e-bikes, there is an increasing need for simulation-based toolsets to evaluate their impact on urban mobility. 
Nowadays, commercial e-mobility platforms such as Bird and Moovit have been widely deployed \cite{lod2024}. However, their proprietary nature imposes significant limitations on researchers and practitioners aiming to explore diverse scenarios and customize solutions. Simulation-based solutions have emerged as a preferable alternative, offering flexibility and adaptability for studying various e-mobility scenarios. In this context, traditional traffic simulation platforms such as AIMSUN \cite{barcelo2005dynamic} and VISSIM \cite{fellendorf2010microscopic} primarily address traffic congestion and scheduling, whereas emerging e-mobility solutions focus on energy management and sustainability. These platforms offer some scope for customization but are still not open-sourced. Among these, the open-source tool Simulation of Urban MObility (SUMO) has gained widespread adoption in the research community due to its versatility and extensibility \cite{krajzewicz2002sumo}. However, despite its strengths, SUMO has certain limitations, particularly in handling large-scale scenarios and integrating with modern cloud-based infrastructures. These limitations become critical in shared e-mobility, where factors such as poor service integration, limited scalability, lack of user-centric solutions, and inadequate energy consumption forecasting remain insufficiently addressed \cite{luo2023fleet}.

To address these challenges, there is a growing need for cloud-based SUMO solutions that leverage the scalability of cloud platforms. Recent advancements in cloud-based SUMO focus on enhancing large-scale simulations and real-time data processing, but most studies concentrate on vehicular network optimization rather than shared e-mobility. \cite{Nguyen2020} offers a high-level cloud-based SUMO framework, but it may be limited to simple queries or purely dashboard-based designs, which are not user-friendly. Meanwhile, the emergence of Large Language Models (LLMs) has opened new opportunities for enhancing intelligent decision-making in traffic-related domains, including delivery scheduling, location prediction, and secure communications. Building on this potential, we propose an LLM-powered, cloud-based platform that integrates dynamic traffic simulation with natural language interaction, aiming to bridge the gap between traditional simulation tools and user-friendly, adaptive shared mobility solutions. In particular, our platform supports scenario-based deployment and evaluation of eHub distributions, addressing the lack of scalable simulation toolsets for shared mobility infrastructures. This work builds upon our previous research on energy modeling \cite{Ding2024, yan2023review}, optimization algorithms \cite{lod2024}, and the multi-modal optimization framework for shared e-mobility \cite{maqsood}. Thanks to the modular and data-centric architecture, our platform can be easily adapted for digital twin modeling and deployment in different regions. It separates the system logic from specific datasets, meaning that with appropriate traffic and mapping information, the platform can recreate realistic local mobility patterns without extensive reengineering. The platform's key contributions are outlined next:
\begin{itemize}
    \item \textbf{A comprehensive, open-source, cloud-based platform for shared e-mobility: }  
    We present a framework for shared e-mobility with modular components for traffic simulation, docking station deployment, multi-modal transport, energy-aware route planning, and user interaction via an Android app using an agent-in-the-loop approach. A Dockerized architecture ensures cloud scalability. Source code is available \footnote{https://github.com/SFIEssential/Essential}.

    \item \textbf{Enhanced querying of simulation and travel data through LLMs:}
    We present the first Retrieval-Augmented Generation (RAG)-based approach tailored for post-simulation querying over structured traffic simulation data and user travel data. By integrating LLMs with schema-level RAG, our platform enables natural language interaction with simulation outputs for different types of users, including end users and system operators. The schema-level RAG framework supports context-aware Text-to-SQL generation, achieving an average execution accuracy of 0.98 on user queries and 0.81 on system operator queries. This empowers both users and system operators to explore personalized, fine-grained insights going beyond predefined charts or views, and makes simulation data more understandable and accessible in real-world scenarios.
    


\end{itemize}

The rest of the paper is structured as follows: Section \ref{sec:litRev} reviews related platforms and compares them with ours. Section \ref{sec:architcture} describes the architecture of the proposed platform. Section \ref{sec:llm} details the implementation of the LLM layer. Section \ref{sec:evaluation} presents the experimental setup and analyzes the results. Finally, Section \ref{sec:conclusions} concludes the paper and outlines directions for future work.


\section{Literature Review} \label{sec:litRev}

\begin{figure*}[ht]
    \centering
    \includegraphics[width=1\linewidth]{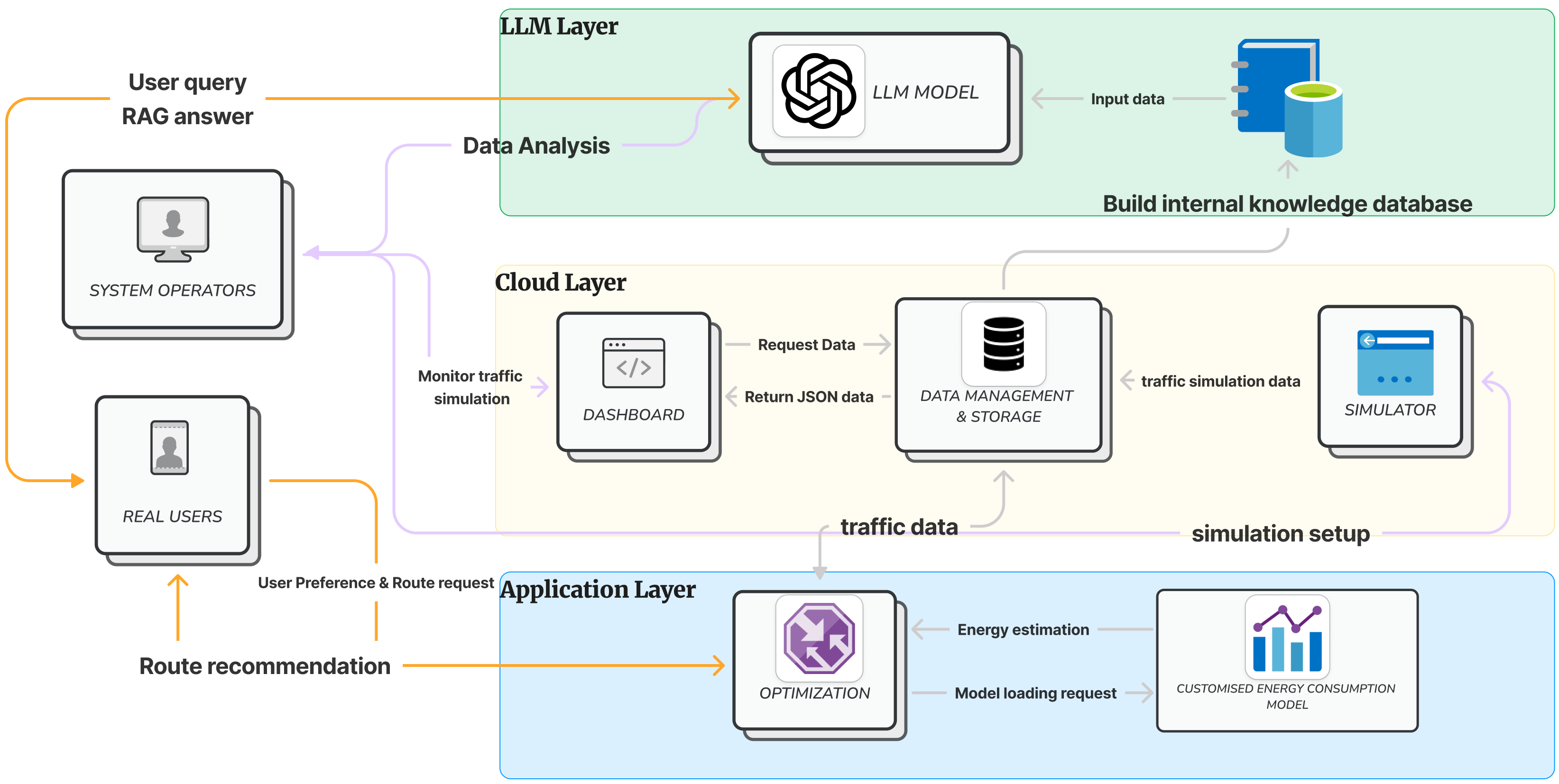}
    \caption{System flow diagram showing the interactions between components.}
    \label{fig:system_flow}
    \vspace{-0.2in}
\end{figure*}

Recent developments in e-mobility platforms aim to tackle urban congestion, environmental concerns, and transport optimization. However, current solutions often lack integration of energy management, real-time cloud simulation, and user-centric features. This section reviews classical traffic simulation platforms, e-mobility platforms, RAG-powered transportation systems, and commercial and public shared e-mobility platforms, highlighting their strengths, limitations, and gaps. It also emphasizes our platform's unique contributions in providing a scalable, multi-modal, and user-focused approach to shared e-mobility simulation and optimization.

\subsection{Transportation Simulation Platform}
Classical traffic simulation platforms, such as AIMSUN~\cite{barcelo2005dynamic} and VISSIM~\cite{fellendorf2010microscopic}, primarily address issues like congestion and scheduling through microscopic modeling~\cite{xiao2005methodology}. SUMO~\cite{krajzewicz2002sumo} further introduced a hybrid approach combining different mathematical models for algorithm evaluation. While traditional platforms focus on congestion and accident analysis, recent efforts have begun integrating e-mobility to assess impacts on energy, urban economies, and communities. Examples include digital twin-based platforms linking e-mobility and smart grids~\cite{cho2022digital}, sustainable EV adoption tools~\cite{ferrara2019simulation}, and data-driven EV impact simulations~\cite{echternacht2018simulating}. Although studies have explored shared e-mobility impacts~\cite{campisi2022promotion, liao2022electric}, few have developed scalable simulation platforms with native support for shared e-mobility and eHub distribution, which is often oversimplified in smart city contexts~\cite{son2022future}. \cite{luo2023fleet} proposed a multi-agent platform using deep reinforcement learning for EV fleet management, but research on scalable, multi-modal shared e-mobility simulation remains limited.

\subsection{LLM-powered Transportation System}

A notable starting point for applying LLMs in transportation is presented in \cite{llm_smart_mobility}, which discusses the utilization of LLMs and RAG in Smart Urban Mobility. However, the study primarily focuses on system design without delving into the implementation details. For traffic scenario generation, the RealGen framework introduced in \cite{realgen_scenario} offers a promising approach. RealGen synthesizes new scenarios by combining behaviors from multiple retrieved examples in a gradient-free manner. These behaviors can originate from predefined templates or tagged scenarios, providing flexibility and adaptability in scenario creation. In the domain of intelligent driving assistance, the IDAS system presented in \cite{idas_rag} demonstrates an efficient approach to reading and comprehending car manuals for immediate, context-based aid. This system provides new insights into enhancing driving assistance by leveraging the capabilities of RAG models. 
Compared to these works, ChatSUMO~\cite{chatsumo} provides a more comprehensive and operational framework by integrating LLMs into the full pipeline of SUMO traffic scenario generation, and customization, significantly reducing the technical barriers for users without traffic simulation expertise. In contrast, our approach focuses specifically on the post-simulation stage, extending LLM integration on the user side to enable schema-level interaction with structured simulation outputs. While prior work primarily emphasizes data generation or simulation modeling, we target the downstream task of querying and interpreting simulation results. Unlike conventional systems that only support fixed dashboard metrics or static visualizations, our platform uniquely integrates RAG-LLM interfaces to allow flexible, natural language querying over arbitrary simulation results.

\section{Cloud-Based Platform Architecture} \label{sec:architcture}
In this section, we present a detailed  overview of the system's architecture and implementation, focusing on its core components and their interconnections.

\subsection{System Overview}
The platform's architecture, shown in Fig. \ref{fig:system_flow}, integrates key components for real-time traffic simulation, data processing, and user interaction. SUMO generates traffic data, which flows through Google Pub/Sub for efficient streaming and into Google Cloud Bigtable (GCB) for storage and analysis. GCB supports real-time visualization via the dashboard and optimal route calculations in the optimization module. Users interact through the dashboard to monitor traffic and manage e-mobility options, while the mobile app provides route recommendations. The schema-level RAG module uses GCB simulation data to generate context-aware responses.


\subsection{Cloud Layer}
\subsubsection{\textbf{Traffic Simulator Module}}
The simulation module is a critical component designed to emulate real-world traffic conditions for evaluating the shared e-mobility framework. It provides a realistic environment to test and validate various aspects, such as traffic patterns, vehicle availability, and user interactions, ensuring that the platform's recommendations are reliable and effective. The simulation process begins with a scenario setup initiated by the system operator, who configures the simulation environment using a JSON file to define key parameters and load configuration data. A central server manages user profiles and processes route requests, forwarding them to a multi-modal optimization module. By leveraging a cloud-based implementation of SUMO, the module ensures scalability and accessibility, making it capable of handling complex, large-scale traffic simulations with efficiency. Pedestrian and bicycle trips in SUMO were generated using \texttt{randomTrips}, with spawn rates varying by period as shown in Table~\ref{tab:route_files}.

\begin{table}[ht]
    \centering
     \caption{Bicycle and Pedestrian Route File Configuration}
    \begin{tabular}{@{}llcc@{}}
        \toprule
        \textbf{Route Type} & \textbf{Simulation Time (seconds)} & \textbf{Spawn Rate (seconds)} \\ 
        \midrule
        \textbf{Bicycles} & 0 - 20,000       & 7            \\ 
                          & 20,000 - 62,000 & 1.5          \\ 
                          & 62,000 - 70,000 & 2.5          \\ 
                          & 70,000 - 86,400 & 5            \\ 
        \midrule
        \textbf{Pedestrians} & 0 - 20,000       & 2            \\ 
                            & 20,000 - 62,000 & 0.75         \\ 
                            & 62,000 - 70,000 & 0.8          \\ 
                            & 70,000 - 86,400 & 1.5          \\ 
        \bottomrule
    \end{tabular}
   
    \label{tab:route_files}
\end{table}



\subsubsection{\textbf{Data Management and Storage Module}}
The data storage module, built on Google Cloud Bigtable (GCB), ensures high-throughput, low-latency storage of time-series traffic data from SUMO simulations and supports the RAG module by providing schema-aware data access. Using SSD-backed storage and Google Cloud Pub/Sub for ingestion, GCB organizes traffic and station data by timestamp and edge ID, while vehicle availability is stored separately. A subscriber script ingests and batches SUMO data efficiently to maintain real-time performance.


\subsubsection{\textbf{Dashboard}}
\begin{figure*}[!htb]
    \centering
    \includegraphics[width=0.9\linewidth]{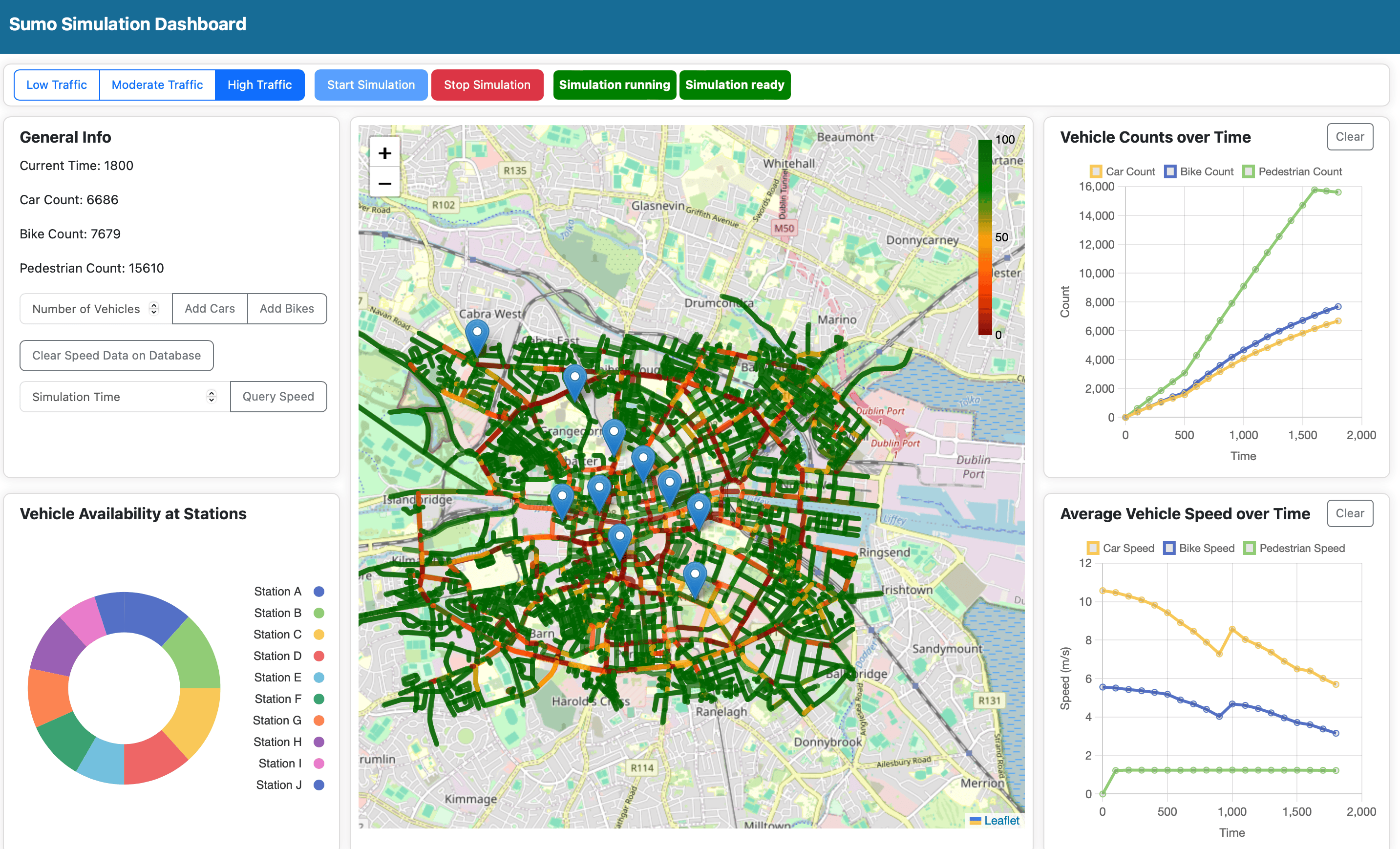}
    \caption{Outlook of the Interactive Dashboard.}
    \label{fig:dashboard_gui}
    \vspace{-0.2in}
\end{figure*}

The interactive dashboard serves as the central interface for controlling and visualizing various aspects of the shared e-mobility emulation platform. It displays real-time traffic simulations for the Dublin City Centre (DCC) map, utilizing GCB as the backend for handling large-scale, time-series traffic data generated by the SUMO system. It ensures high performance and scalability, offering users a seamless way to set up scenarios, control simulations, and analyze traffic patterns in real-time, as illustrated in Fig. \ref{fig:dashboard_gui}.

\subsection{Application Layer}
\subsubsection{\textbf{Optimization Module}}
The multimodal transportation optimization module is designed to provide user-centric solutions while leveraging a strategically pre-designed distribution of transportation stations. This framework integrates real-time traffic simulation data from SUMO with user-specific inputs such as origins, destinations, and transportation preferences. At the core of the framework is a Mixed-Integer Linear Programming (MILP) model, which optimizes travel time while adhering to constraints such as energy consumption and maximum transfer times. and support multi-modal e-mobility transportation. The detailed solution has been given in previous work \cite{lod2024}. The energy consumption is predicted by energy model in \cite{Ding2024}. We employ the user-in-the-loop design principle to allow users to visualize and interact with optimal multi-modal routes under different scenario configurations, such as varying eHub distributions or traffic conditions. Unlike traditional static routing systems, our approach enables users to explore dynamic trade-offs between travel time, energy consumption, and transfer flexibility in real-time, fostering a deeper understanding of sustainable mobility choices.

\subsubsection{\textbf{Android Mobile Application}}
\begin{figure}[htbp]
    \centering
    \begin{subfigure}{0.2\textwidth}
        \centering
        \includegraphics[width=0.9\linewidth, height=6cm, keepaspectratio]{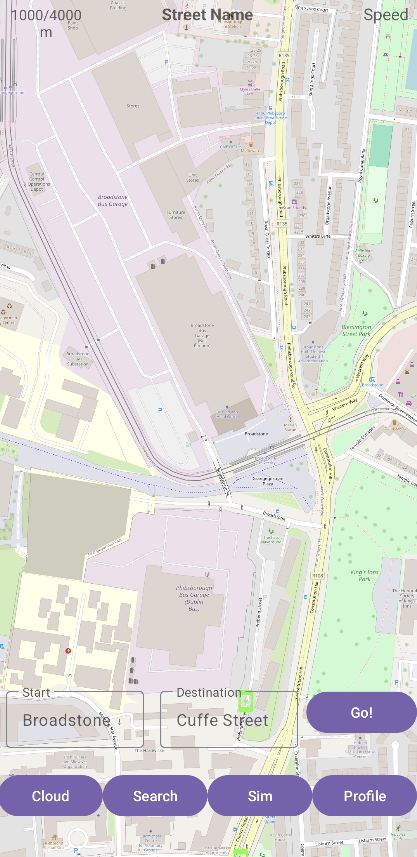}
        \caption{User input of start and destination locations.}
        \label{fig:app-input}
    \end{subfigure}
    \hspace{0.01\textwidth}
    \begin{subfigure}{0.2\textwidth}
        \centering
        \includegraphics[width=\linewidth, height=6cm, keepaspectratio]{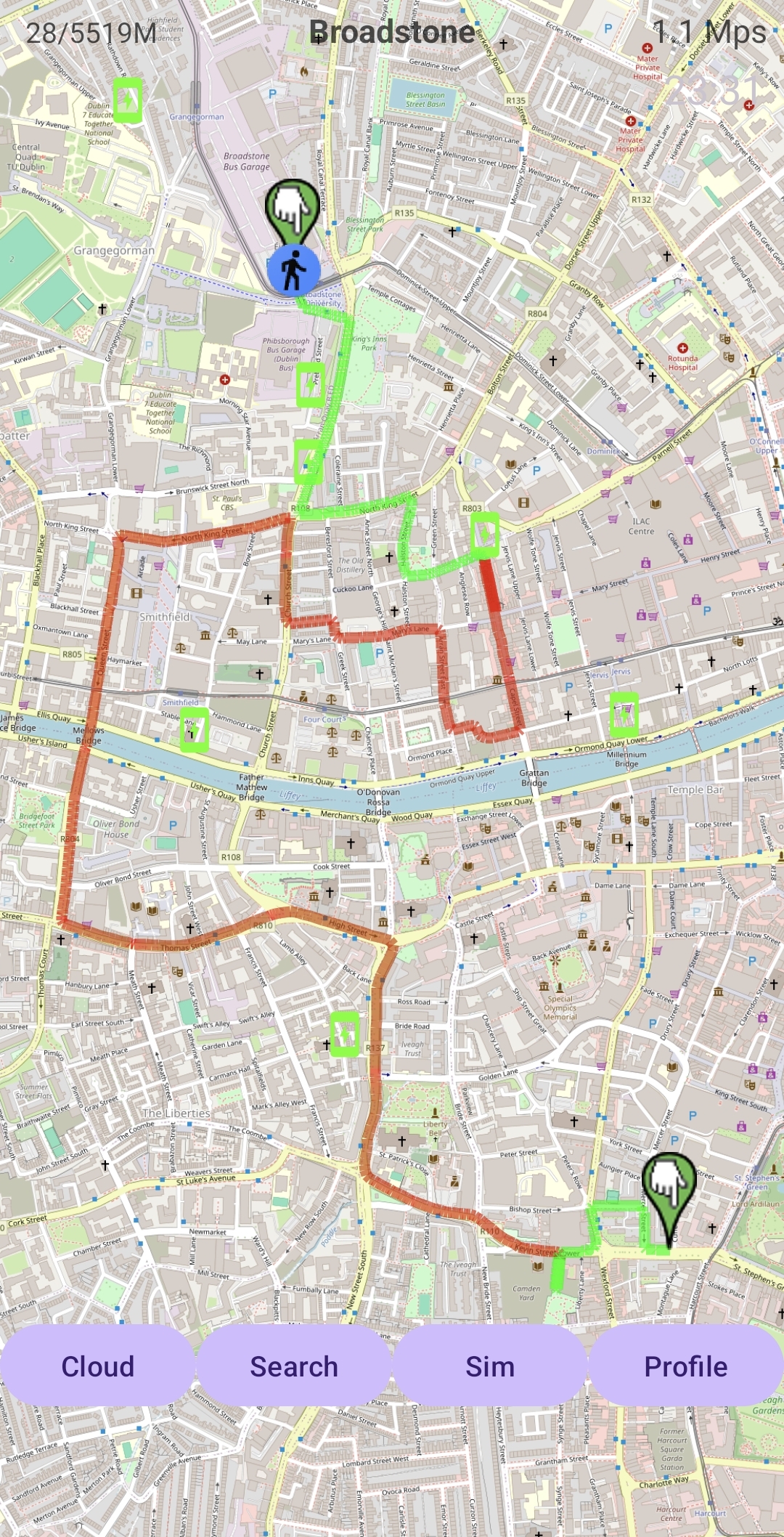}
        \caption{Route with different transportation modes.}
        \label{fig:app-route}
    \end{subfigure}
    \caption{Screenshots of the Android app: users select start and destination points (a), and the app recommends a route (b), with green for walking and red for e-scooter paths.}
    \label{fig:app-demo}
    \vspace{-0.3in}
\end{figure}

An Android app has been developed to improve e-mobility route planning. Users input their origin, destination, and trip preferences. Leveraging real-time traffic simulation data, the app generates optimized routes based on time and energy constraints. Routes are visualized dynamically with color-coded transport modes and estimated speed. Fig.~\ref{fig:app-input} and Fig.~\ref{fig:app-route} illustrate the input and route display features. The app offers an efficient travel solution with real-time data integration.

\section{LLM Layer: RAG Module} \label{sec:llm}
We introduce a two-stage Text-to-SQL generation framework that integrates schema-aware retrieval with LLMs. Our system supports two types of structured databases: (i) system-level traffic simulation data and (ii) user-level historical travel records. We assume two types of users in the system: system operators, who are responsible for managing and monitoring simulation data; and mobile users, who interact with the system based on their travel history. The implementations steps are as follows:
\subsubsection{\textbf{M-Schema Generation}}
M-Schema is proposed by the XiYanSQL project,\footnote{\url{https://github.com/XGenerationLab/M-Schema}} is a structured format that encodes relational database schemas into a model-friendly JSON, preserving both table- and column-level semantics. Compared to unstructured table documentation, it better supports LLMs in aligning natural language queries with database structure. We adopt the official open-source M-Schema framework, which enables automatic conversion from relational databases. To generate M-Schema from our PostgreSQL database which is linked with GCB, we follow these steps:
\begin{enumerate}
    \item Connect to the database: We use SQLAlchemy to connect to the PostgreSQL instance.
    \item Parse schema: The \texttt{schema\_engine.py} script introspects the database schema and extracts metadata including table names, primary/foreign key relationships, and column types.
    \item Generate M-Schema JSON: The extracted schema is transformed into a JSON file containing structured representations for each table and its columns.
\end{enumerate}


\subsubsection{\textbf{Schema Embedding and Storage}}

To support schema-aware generation, we first convert all database M-Schema descriptions into vector representations. Each schema is written as a separate natural language text file and loaded using a preprocessing script. We use the \texttt{all-MiniLM-L6-v2} model from SentenceTransformers to encode these schema texts into dense embeddings. The resulting vectors are stored in a local Chroma vector database, where each schema document is indexed with a unique ID and associated metadata. 

\subsubsection{\textbf{Schema Indexing}}

At query time, a user-provided natural language question is encoded into the same embedding space using the same transformer model. A vector similarity search is then performed against the stored schema embeddings using cosine similarity, defined as:

\begin{align}
\mathit{sim}(\boldsymbol{q}, \boldsymbol{d}) 
&= \frac{\boldsymbol{q} \cdot \boldsymbol{d}}{\|\boldsymbol{q}\| \, \|\boldsymbol{d}\|} \label{eq:cosine}
\end{align}

where $\boldsymbol{q}$ and $\boldsymbol{d}$ represent the embedding vectors of the query and a schema document, respectively.


We retrieve the top-3 most semantically relevant schema documents based on their similarity scores. These documents are then concatenated to form the context input for LLMs. This retrieval mechanism supplies schema-specific context prior to SQL generation.


\subsubsection{\textbf{SQL Generation}}
We adopt five different LLMs to generate SQL queries given a natural language question and the retrieved schema context:

    
    



\begin{itemize}
    \item \textbf{XiYanSQL:} A schema-aware Text-to-SQL model using M-Schema as input, selected for its top-three ranking on BirdSQL\footnote{\url{https://bird-bench.github.io/}} and its publicly available API.
    
    \item \textbf{GPT-4:} OpenAI’s general-purpose LLM with strong Text-to-SQL performance, including on BirdSQL.

    \item \textbf{GPT-4o:} A low-latency, cost-efficient variant of GPT-4 that maintains competitive Text-to-SQL performance and ranks highly on BirdSQL.

    \item \textbf{Gemini 1.5 Pro:} Google’s multimodal model excelling in long-context and Text-to-SQL tasks.

    \item \textbf{Gemini 2.0 Flash:} A faster, enhanced Gemini variant optimized for everyday tasks.
\end{itemize}

\section{Experiments and Evaluation} \label{sec:evaluation}

In this section, we introduce the experiments which are conducted to evaluate the platform and analyze the results.

\subsection{Experiment Setup}
\subsubsection{Simulation Setup}

Our simulation was based on the DCC map, chosen for its links to prior research in this field \cite{gueriau2020quantifying}. The map, covering a 5 km x 3.5 km area, provided a robust foundation for simulating a 24-hour workday, using vehicle counts collected every 6 minutes from 480 locations via Dublin's SCATS system. Data from Transport Infrastructure Ireland, averaged over several months, modeled realistic traffic behavior. To support multi-modal simulations, modifications were made to include pedestrian and bicycle lanes using SUMO's \texttt{Netconvert} tool. Traffic data from the 2022 Irish census \cite{cso2023census}, showing 87,000 walkers and 37,000 cyclists, were incorporated, with peak times between 5:30 am and 7 pm.

\subsubsection{Optimization Module Evaluation Setup}
The evaluation of our optimization model was conducted using simulated data generated from a traffic simulator. The traffic scenarios considered include \textit{Low Traffic}, \textit{Medium Traffic}, and \textit{High Traffic}, reflecting varying congestion levels within an urban environment. The transportation network was modeled using the road network provided in the DCC.net.xml file, which was parsed and transformed into a directed graph using the NetworkX library.
The optimization problem was formulated and solved using the PuLP library with the CBC solver, ensuring efficient computation of the optimal routes for different traffic conditions. The setup involved generating paths for given 400 Origin-Destination (OD) pairs. Vehicle types considered include e-cars, e-bikes, and e-scooters, along with walking as an option. The performance of the module was assessed by analyzing the extra time costs under different traffic conditions.

\subsubsection{RAG Module Evaluation Setup}
The Schema-Level RAG module was evaluated from two perspectives: system operators and mobile users. For system operators, the retrieved data consisted of traffic simulation outputs. For mobile users, the queried data included each user's trip history stored on cloud. The evaluation for the Schema-Level RAG module was implemented in Python with the following key components:

\noindent \textbf{Traffic Data:} A database containing edge-level traffic information such as edge\_id, simulation\_time, pedestrian\_speed, bike\_speed, and car\_speed, as well as station-level data including vehicle types and battery levels.

\noindent \textbf{User Travel History:} Simulated user-level trajectory data, including start\_edge, end\_edge, travel time cost, execution time, and optimal path sequence.

\noindent \textbf{Evaluation Queries:} A set of 100 natural language questions and corresponding ground-truth SQL queries were manually created for each type of user, resulting in 200 annotated QA pairs. These ensure high-quality evaluation and reliable SQL correctness checking. 


\noindent \textbf{Large Language Models:} XiYanSQL, GPT-4, GPT-4o, Gemini 1.5 Pro, Gemini 2.0 Flash.

\noindent \textbf{Embedding Model:} \texttt{all-MiniLM-L6-v2} for similarity-based retrieval and question similarity scoring.

\noindent \textbf{Evaluation Metrics:} For structured SQL generation evaluation, we follow the Spider benchmark~\cite{yu2018spider}:
\begin{itemize}
    \item \textbf{Execution Accuracy:} Whether executing the predicted SQL matches the ground-truth result.
    \item \textbf{Component Match (F1):} Token overlap across SELECT, WHERE, GROUP BY, and ORDER BY clauses.
    \item \textbf{BLEU Scores (1–4):} N-gram overlaps with smoothing to measure surface similarity.
    \item \textbf{ROUGE Scores:} ROUGE-1, ROUGE-2, and ROUGE-L for assessing overlap and fluency.
\end{itemize}

\begin{figure}[!htb]
    \centering
    \includegraphics[width=1\linewidth]{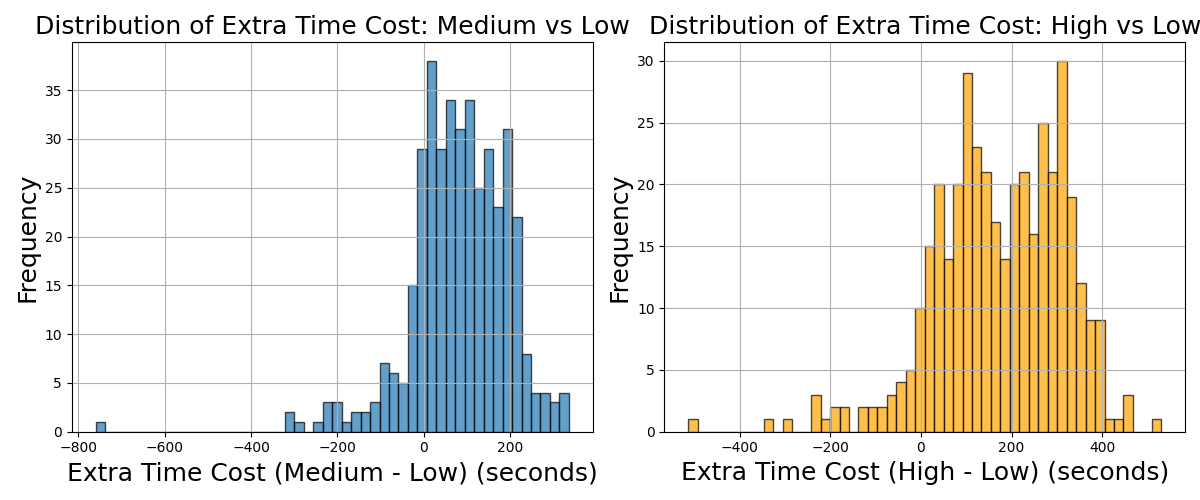}
    \caption{Distribution of extra travel time cost under different traffic scenarios.}
    \label{fig:histogram_plot}
\end{figure}
\subsection{Evaluation and Analysis}
\begin{table*}[!b]
\centering
\scriptsize
\begin{tabular}{llcccccccccc}
\toprule
\textbf{Model} & \textbf{Domain} & \textbf{Execution Accuracy} & \textbf{F1} & \textbf{BLEU-1} & \textbf{BLEU-2} & \textbf{BLEU-3} & \textbf{BLEU-4} & \textbf{ROUGE-1} & \textbf{ROUGE-2} & \textbf{ROUGE-L} \\
\midrule
XiYanSQL       & System & \textbf{0.81}   & \textbf{0.9183} & \textbf{0.8360} & \textbf{0.7825} & \textbf{0.7212} & 0.6822 & \textbf{0.8916} & \textbf{0.7802} & \textbf{0.8867} \\
GPT-4          & System & 0.74            & 0.8870          & 0.8101          & 0.7647          & 0.7238          & \textbf{0.6959} & 0.8753          & 0.7594          & 0.8710 \\
GPT-4o         & System & 0.70            & 0.8814          & 0.7880          & 0.7291          & 0.6664          & 0.6272 & 0.8664          & 0.7297          & 0.8637 \\
Gemini 2.0 Flash & System & 0.66          & 0.8746          & 0.7026          & 0.6010          & 0.5224          & 0.4718 & 0.7792          & 0.5633          & 0.7703 \\
Gemini 1.5 Pro & System & 0.65            & 0.8613          & 0.7594          & 0.7083          & 0.6705          & 0.6433 & 0.8375          & 0.7052          & 0.8319 \\

\midrule
XiYanSQL       & User   & \textbf{0.98}   & \textbf{0.9755} & \textbf{0.9526} & \textbf{0.9371} & \textbf{0.9125} & \textbf{0.8999} & \textbf{0.9698} & \textbf{0.9358} & \textbf{0.9698} \\
GPT-4          & User   & 0.89            & 0.9119          & 0.9330          & 0.9096          & 0.8889          & 0.8765 & 0.9483          & 0.9014          & 0.9467 \\
Gemini 2.0 Flash & User & 0.88            & 0.9061          & 0.9213          & 0.8902          & 0.8694          & 0.8512 & 0.9390          & 0.9126          & 0.9529 \\
GPT-4o         & User   & 0.84            & 0.8981          & 0.9052          & 0.8721          & 0.8380          & 0.8185 & 0.9329          & 0.8671          & 0.9321 \\
Gemini 1.5 Pro & User & 0.83          & 0.8526          & 0.8439         & 0.8172          & 0.7275          & 0.7192 & 0.8590          & 0.8267          & 0.8973 \\
\bottomrule
\end{tabular}
\caption{Evaluation of SQL generation on system-level and user-level queries. All scores are reported as decimals.}
\label{tab:model-eval}
\vspace{-0.15in}
\end{table*}

\begin{figure*}[!t]
    \centering
    \includegraphics[width=1\textwidth]{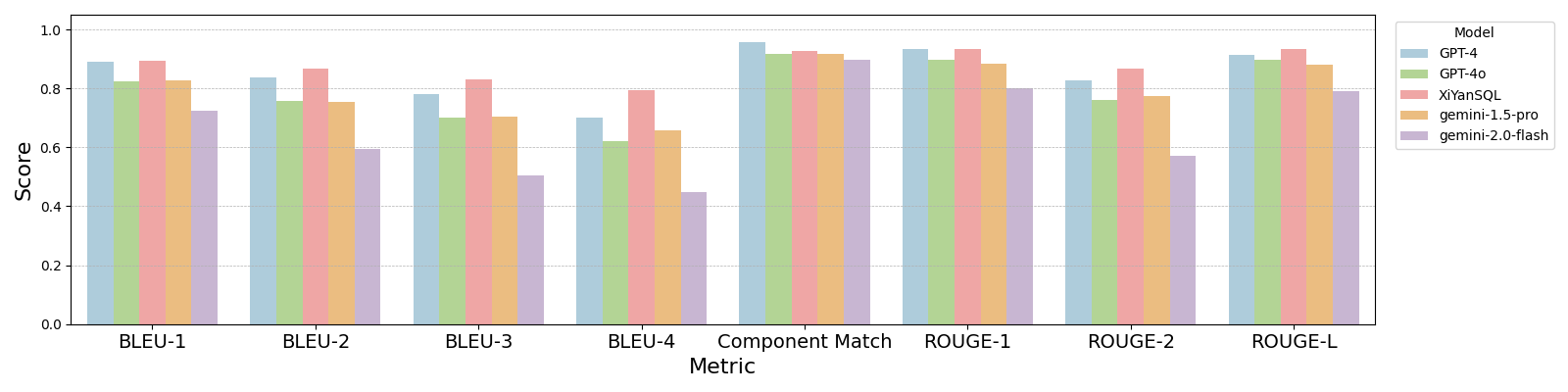}
    \vspace{-0.3in}
    \caption{Comparison of Median Value of Evaluation Metrics for Models on System Operator Queries.}
    \label{fig:barchart}
    \vspace{-0.1in}
\end{figure*}

\begin{figure*}[!htb]
    \centering
    \includegraphics[width=1\textwidth]{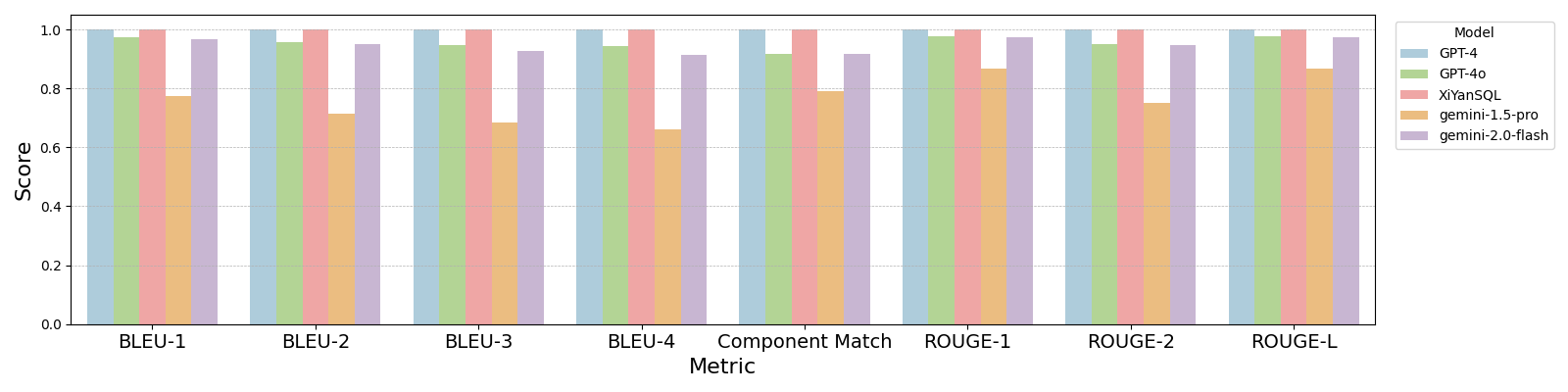}
    \vspace{-0.3in}
    \caption{Comparison of Median Value of Evaluation Metrics for Models on User Queries.}
    \label{fig:bar-user}
    \vspace{-0.1in}
\end{figure*}

\subsubsection{Optimization Module}

The results of the optimization module evaluation are presented in Fig. \ref{fig:histogram_plot}, which compares the extra travel time cost required for all OD pairs across different traffic scenarios. The figure displays two histograms: one comparing medium vs. low traffic scenarios and another comparing high vs. low traffic scenarios. These distributions provide insights into the additional delays introduced by increasing congestion levels.

The key observations from the results are as follows:

\begin{itemize}
    \item \textbf{Medium vs. Low Traffic:} The histogram shows a moderate increase in extra travel time with a wider spread compared to the low traffic scenario, and the Extra Time Cost (Medium - Low) expectation is 81.01 seconds. This suggests that congestion begins to impact route efficiency, but remains manageable.
    \item \textbf{High vs. Low Traffic:} A significant increase in travel time cost is observed, with a much broader distribution, and the Extra Time Cost (High - Low) expectation is 172.42 seconds. This indicates higher variability and suggests that congestion has a substantial effect on route optimization, leading to more frequent delays.
\end{itemize}

\subsubsection{RAG module}

For schema-level RAG, as shown in Table~\ref{tab:model-eval}, XiYanSQL achieves the highest execution accuracy with 0.81 on system operator queries and 0.98 on user-level queries. GPT-4 obtains the highest BLEU-4 score of 0.6959 on system operator queries. Overall, models perform better on user queries due to simpler schema structures. Fig.~\ref{fig:barchart} illustrates the median scores across multiple evaluation metrics for each model on system operator queries. GPT-4 and XiYanSQL consistently perform well across BLEU and ROUGE metrics, with GPT-4 showing the highest BLEU-1 to BLEU-4 scores and strong component match accuracy. Fig.~\ref{fig:bar-user} presents the median scores across different evaluation metrics on user queries. XiYanSQL maintains consistently high median values in all metrics including BLEU, Component Match and ROUGE, indicating both accuracy and stability in its outputs. GPT-4 and GPT-4o also perform well, though slight drops appear in BLEU-3 and BLEU-4. In contrast, the Gemini models, especially gemini-1.5-pro, have lower median scores in BLEU-3, BLEU-4 and ROUGE-2, reflecting difficulties in generating coherent and complete responses. These trends support that simpler schemas in user queries yield better and more consistent model performance.

\begin{figure*}[h]
    \centering
    \includegraphics[width=1\textwidth,height=6cm]{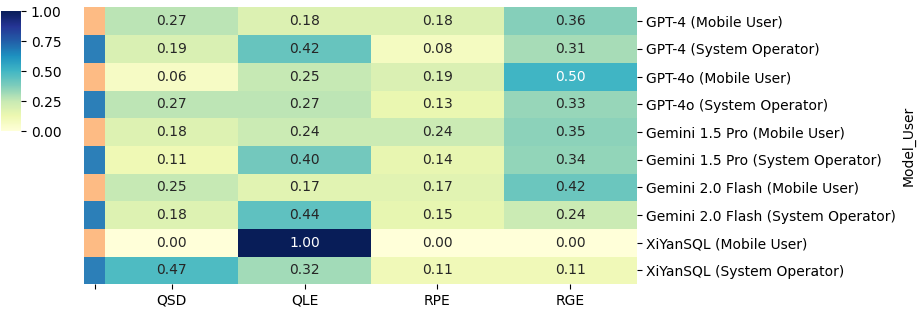}
    \caption{Error Type Proportions Across Different Models and Users.}
    \label{fig:heatmap}
    \vspace{-0.1in}
\end{figure*}

Furthermore, we define four error types to analyze model performance: Query Structure Differences (QSD), including mismatched fields, incorrect use of LIMIT, wrong table choices, and aggregation mistakes; Query Logic Errors (QLE), such as incomplete or incorrect JOIN and filter conditions; Result Precision Errors (RPE), caused by missing operators like DISTINCT or ABS; and Result Granularity Errors (RGE), related to selecting too many or too few fields. This taxonomy extends \cite{classification} by distinguishing precision from granularity issues. As visualized in Fig.~\ref{fig:heatmap}, each cell represents the proportion of a specific error type normalized by the total number of errors for each model-user pair. XiYanSQL shows a high rate of structure-related errors on system operator queries, with QSD accounting for 47\% of its total. GPT-4 and GPT-4o exhibit the highest share of logic errors, at 42\% and 44\% respectively, especially under complex conditions. Gemini models show more than 30\% granularity errors on user queries, indicating difficulty in selecting the appropriate number of fields. Precision errors remain consistently low across all models, generally below 20\%. These results suggest that schema alignment is model-specific, while logic and granularity issues are common across schema-level RAG.

Table~\ref{tab:correct} and Table~\ref{tab:incorrect} present representative examples of correct and incorrect SQL generated by XiYanSQL. The correct cases demonstrate the model’s ability to capture the core intent and structure of the query, even when using alternative but semantically equivalent SQL formulations. In contrast, the incorrect cases reveal common failure patterns, such as mismatched aggregation conditions, which often lead to semantically inaccurate results despite syntactic similarity.

\begin{table*}[ht]
\centering

\begin{tabularx}{\textwidth}{@{}lX@{}}
\toprule
\textbf{System Operator Question:} & Which edge\_id has the most stations? \\
\textbf{Ground Truth SQL:} & SELECT edge\_id, COUNT(*) AS station\_count FROM stations GROUP BY edge\_id ORDER BY station\_count DESC LIMIT 1; \\
\textbf{Generated SQL:} & SELECT edge\_id, COUNT(station\_id) AS station\_count FROM stations GROUP BY edge\_id ORDER BY station\_count DESC LIMIT 1; \\
\midrule

\textbf{User Question:} & Find the top 3 most frequent destinations. \\
\textbf{Ground Truth SQL:} & SELECT end\_edge, COUNT(*) AS freq FROM user\_paths GROUP BY end\_edge ORDER BY freq DESC LIMIT 3; \\
\textbf{Generated SQL:} & SELECT end\_edge, COUNT(*) AS end\_edge\_count FROM user\_paths GROUP BY end\_edge ORDER BY end\_edge\_count DESC LIMIT 3; \\
\bottomrule
\end{tabularx}
\caption{Correct SQL Predictions from XiYanSQL}
\label{tab:correct}

\end{table*}

\begin{table*}[ht]
\centering

\begin{tabularx}{\textwidth}{@{}lX@{}}
\toprule
\textbf{System Operator Question:} & Get the average bike speed for the road segments that have at least one station. \\
\textbf{Ground Truth SQL:} & SELECT o.edge\_id, AVG(o.bike\_speed) AS avg\_bike\_speed FROM online\_demo o JOIN stations s ON o.edge\_id = s.edge\_id GROUP BY o.edge\_id; \\
\textbf{Generated SQL:} & SELECT AVG(bike\_speed) AS average\_bike\_speed FROM online\_demo WHERE edge\_id IN (SELECT edge\_id FROM stations); \\
\midrule

\textbf{User Question:} & Find paths where the optimal route contains more than 4 road segments. \\
\textbf{Ground Truth SQL:} & SELECT * FROM user\_paths WHERE LENGTH(optimal\_path\_sequence) - LENGTH(REPLACE(optimal\_path\_sequence, '(', '')) \textgreater{} 4; \\
\textbf{Generated SQL:} & SELECT * FROM user\_paths WHERE LENGTH(optimal\_path\_sequence) - LENGTH(REPLACE(optimal\_path\_sequence, ',', '')) \textgreater{} 4; \\
\bottomrule
\end{tabularx}
\caption{Incorrect SQL Predictions from XiYanSQL}
\label{tab:incorrect}

\vspace{-0.2in}
\end{table*}

\section{Conclusion and Future Works}\label{sec:conclusions}

This paper presents a comprehensive shared e-mobility platform that integrates cloud-based simulation, real-time data processing, and RAG-enhanced decision-making to address the challenges in sustainable urban transportation. The platform supports dynamic multi-modal routing and flexible docking-based systems through a scalable and user-centric design. The schema-level RAG evaluation highlights XiYanSQL's high execution accuracy of 0.81 on system operator queries and 0.98 on user queries. An error analysis reveals that logic errors are the predominant failure type for most models, while granularity issues are more common in user-level queries. Although the LLM-based RAG framework performs robustly, the generated responses are not always perfect. Users and system operators should treat LLM outputs as supportive decision-making aids rather than definitive conclusions and should cross-check critical decisions against simulation data or dashboard outputs when high reliability is required.  

Finally, some limitations remain to be addressed in this work. First, the platform’s scalability under high user loads and real-time traffic updates has not been fully assessed. Second, its adaptability to different urban networks and traffic configurations requires further investigation. Third, usability testing has yet to be conducted. As part of the future work, we will address these gaps by focusing on large-scale performance, cross-platform portability, integration with public transportation, and usability studies with a view to further enhance user interaction and system responsiveness.

\newpage

\bibliographystyle{IEEEtran}
\bibliography{reference}

\end{document}